\def\year{2021}\relax
\documentclass[letterpaper]{article} 
\usepackage{aaai21}  
\usepackage{times}  
\usepackage{helvet} 
\usepackage{courier}  
\usepackage[hyphens]{url}  
\usepackage{graphicx} 
\usepackage{graphicx}  
\usepackage{natbib}  
\usepackage{caption} 
\usepackage{booktabs}
\usepackage{csquotes}
\usepackage{amsmath}
\usepackage{tabularx}
\usepackage[linesnumbered,ruled, algo2e]{algorithm2e}
\usepackage{xcolor}

\SetCommentSty{mycommfont}
\SetKwComment{Comment}{$\triangleright$\ }{}
\usepackage{multirow}

\title{On Generating Plausible Counterfactual and \\ Semi-Factual Explanations for Deep Learning}
\author{
    Eoin M. Kenny* and Mark T. Keane
}
\affiliations{
\\ 
    University College Dublin, Dublin, Ireland\\
    Insight Centre for Data Analytics, UCD, Dublin, Ireland \\
    VistaMilk SFI Research Centre \\
    \texttt{eoin.kenny@insight-centre.org, mark.keane@ucd.ie}
}

\begin{document}

\maketitle

\begin{abstract}
There is a growing concern that the recent progress made in AI, especially regarding the predictive competence of deep learning models, will be undermined by a failure to properly explain their operation and outputs. In response to this disquiet, counterfactual explanations have become massively popular in eXplainable AI (XAI) due to their proposed computational, psychological, and legal benefits. In contrast however, semi-factuals, which are a similar way humans commonly explain their reasoning, have surprisingly received no attention. Most counterfactual methods address tabular rather than image data, partly due to the latter's non-discrete nature making good counterfactuals difficult to define. Additionally, generating plausible looking explanations which lie on the data manifold is another issue which hampers progress. This paper advances a novel method for generating plausible counterfactuals (and semi-factuals) for black-box CNN classifiers doing computer vision. The present method, called PlausIble Exceptionality-based Contrastive Explanations (PIECE), modifies all \enquote{exceptional} features in a test image to be \enquote{normal} from the perspective of the counterfactual class (hence concretely defining a counterfactual). Two controlled experiments compare this method to others in the literature, showing that PIECE not only generates the most plausible counterfactuals on several measures, but also the best semi-factuals.
\end{abstract}


\section{Introduction}
In the last few years, emerging issues around the the \textit{interpretability} of machine learning models have elicited a major, on-going response from government~\cite{gunning2017explainable}, industry~\cite{pichai_2018}, and academia~\cite{miller2019explanation} on eXplainable AI (XAI)~\cite{guidotti2018survey,adadi2018peeking}. As opaque, black-box deep learning models are increasingly being used in the \enquote{real world} for high-stakes decision making (e.g., medicine and law), there is a pressing need to give end-users some insight into how these models achieve their predictions. In this paper, we advance a new technique for XAI using counterfactual and semi-factual explanations, applied to deep learning models [i.e., convolutional neural networks (CNNs)]. These \enquote{contrastive explanations} have attracted massive interest in AI~\cite{miller2018contrastive,wachter2017counterfactual}, but have never directly examined semi-factual explanations. This is important because counterfactual explanations appear to offer computational, psychological, and legal advantages over other explanation strategies, and semi-factuals should also. In this introduction, we review the importance of contrastive explanation and related work.

\subsection{Contrastive Explanation}

\begin{figure}[!t]

  \centering
  \includegraphics[width=\columnwidth]{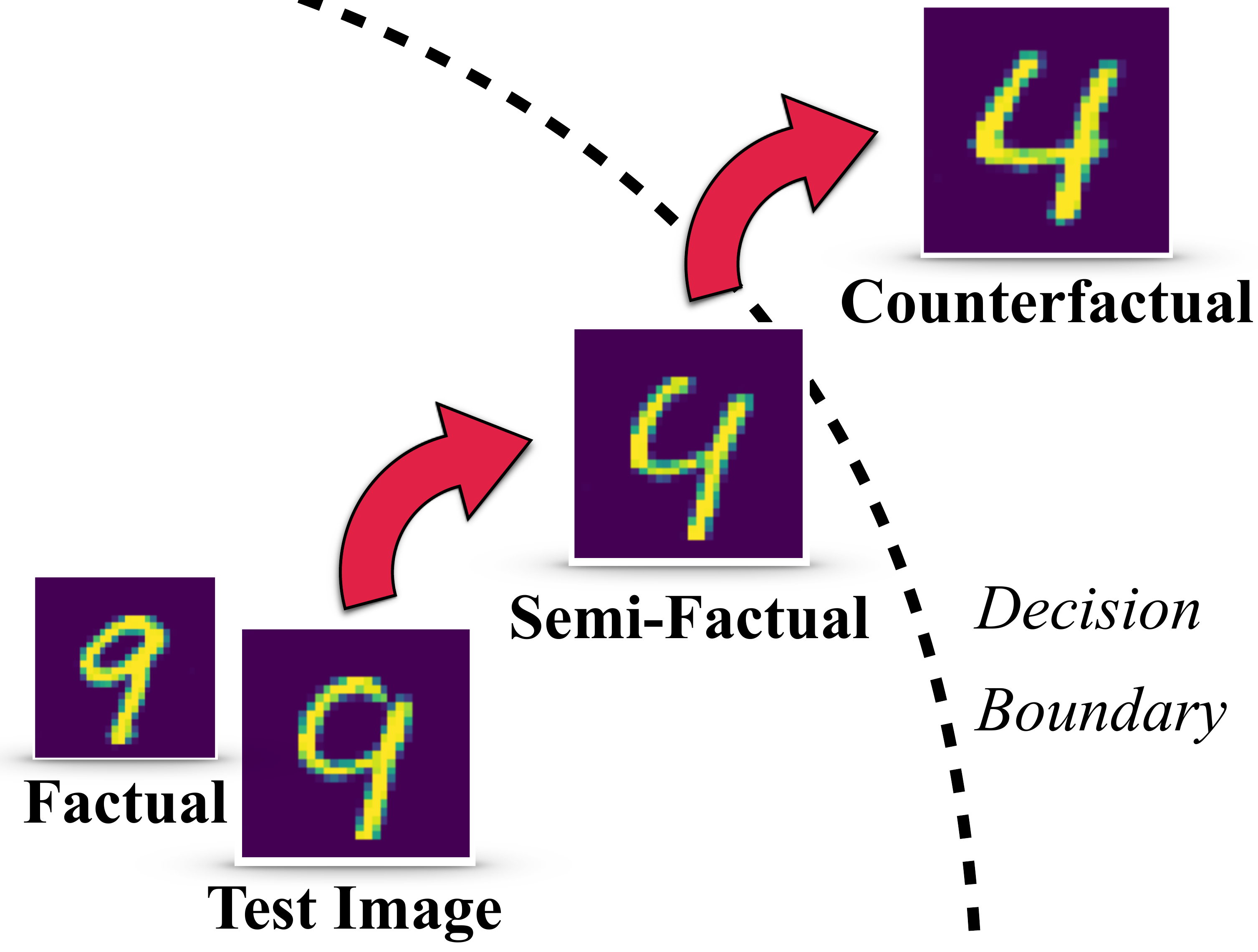} 
  \caption{PIECE: A test image is shown next to a nearest neighbor from the training data (i.e., a factual explanation simply for context here), alongside a synthetic semi-factual and counterfactual explanation generated for the test image by PIECE.}
  \label{Fig:1}
\end{figure}

To understand what makes counterfactuals important, consider the difference between factual and counterfactual explanations. An AI loan application system could explain its decision \textit{factually} saying \enquote{You were refused because a previous customer with your profile asked for this amount, and was also refused}. In contrast, a \textit{counterfactual} explanation of the same refusal might say \enquote{If you applied for a slightly lower amount, you would have been accepted}. The proponents of counterfactuals argue that they have distinct computational, psychological, and legal benefits for XAI. Computationally, counterfactuals provide explanations without having to \enquote{open the black box}~\cite{grath2018interpretable}. Psychologically, counterfactuals elicit spontaneous, causal thinking in people, thus making explanations that use them more engaging~\cite{byrne2019counterfactuals,miller2019explanation}. Legally, it is argued that counterfactual explanations are GDPR compliant~\cite{wachter2017counterfactual}.

\begin{figure*}[!t]

  \centering
  \includegraphics[width=\textwidth]{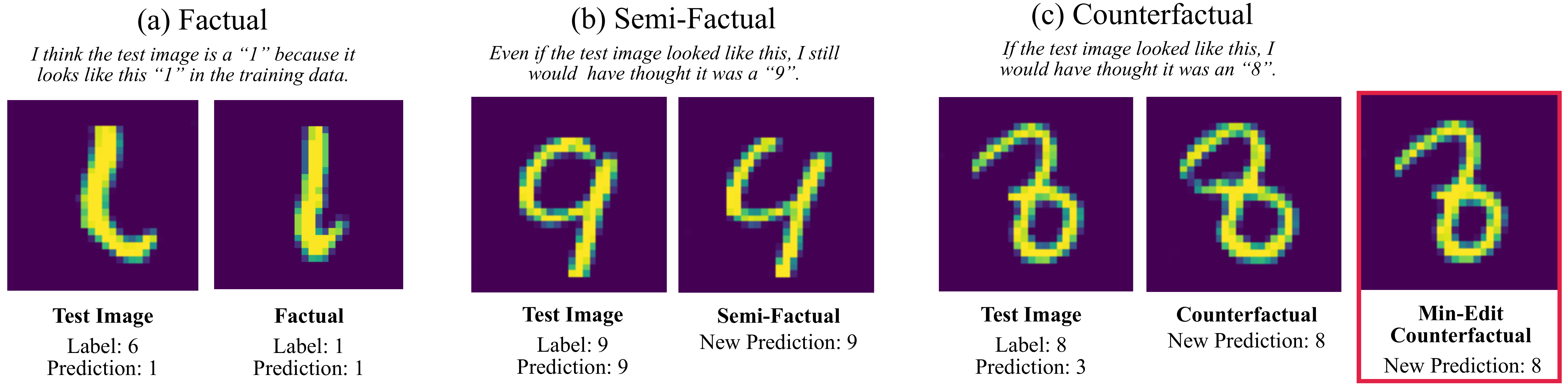} 
  \caption{\textit{Post-Hoc} Factual, Semi-Factual, and Counterfactual Explanations on MNIST: (a) a \textit{factual explanation} for a misclassification of \enquote{6} as \enquote{1}, that uses a nearest-neighbor in latent-space classed as \enquote{1}, (b) a \textit{semi-factual explanation} for the correct classification of a \enquote{9}, that shows a synthetic instance with meaningful feature changes that would \textit{not} alter its classification, and (c) a \textit{counterfactual explanation} for the misclassification of an \enquote{8} as a \enquote{3}, that shows a synthetic instance with meaningful feature changes that would cause the CNN to correct its classification (n.b., for comparison a counterfactual using the \textit{Min-Edit} method (see Expt.~1) is shown with its human-undetectable feature-changes).}
  \label{fig:intro}
  \label{Fig:2}
\end{figure*}

Similar arguments for counterfactuals can be also made for semi-factual explanations, were humans typically begin an explanation with the words \enquote{\textit{Even if}...}. For example, the previous AI loan system might say \enquote{\textit{Even if} you had asked for a slightly lower amount, you still would have been refused}. Semi-factuals are a common form of human explanation and have been researched in psychology for decades~\cite{mccloy2002semifactual}, they offer the benefits of contrastive explanations (e.g., counterfactuals) without having to cross a decision boundary, which in turn decreases the amount of featural changes needed to convey an explanation. This is important because the less featural changes there are, the more interpretable the explanation likely is~\cite{keanesmyth2020}. This issue is one of the main drawbacks of counterfactual explanations, which semi-factuals can conceivably help correct. Despite this however, semi-factual reasoning has been largely ignored in the AI community, they sit between factuals and counterfactuals (see Fig.~\ref{Fig:1}), offering causal justifications for same-class predictions. Additionally, semi-factuals have the advantage of decreasing negative emotions in people when compared to counterfactuals~\cite{mccloy2002semifactual}, which may have a notable use when giving explanations for bad news such as a loan rejection, or a devastating medical diagnosis. Lastly, semi-factuals can make a prediction seem incontestable~\cite{byrne2019counterfactuals}, which is highly effective for convincing people a classifier is correct~\cite{nugent2009gaining}. 

These explanation strategies for interpreting AI models -- factual, counterfactual, and semi-factual -- are typically used for \textit{post-hoc} explanation-by-example ~\cite{lipton2018mythos}. In general, \textit{post-hoc} explanations provide after-the-fact justifications for why a prediction was made using nearest-neighbor training instances~\cite{kenny2019twin}, generated synthetic instances~\cite{wachter2017counterfactual}, or feature contributions~\cite{ribeiro2016should}.

\subsection{Related Work} 
Most \textit{post-hoc} explanation-by-example research on counterfactuals has focused on discrete data such as tabular datasets [e.g., see~\cite{grath2018interpretable}]. These methods aim to generate minimally-different counterfactual instances that can plausibly explain test instances [i.e., instances from a \enquote{possible world}~\cite{pawelczyk2020learning}].\footnote{There is a literature using Causal Bayesian Networks to assess fairness of AI systems~\cite{pearl2000causality}. This is a different use of counterfactuals for another aspect of XAI.} These counterfactual explanation techniques can be divided into \enquote{blind perturbation} and \enquote{experience-guided} methods~\cite{keanesmyth2020}. \textit{Blind perturbation} methods generate candidate counterfactual explanations by perturbing feature values of the test instance to find minimally-different instances from a different/opposing class [e.g., \cite{wachter2017counterfactual}], using distance metrics to select \enquote{close} instances. \textit{Experience-guided} methods rely more directly on the training data by justifying counterfactual selection using training instances~\cite{laugel2019dangers}, analyzing features of the training data~\cite{grath2018interpretable}, or by directly adapting training instances~\cite{keanesmyth2020}. At present, it is unclear which works best, as there is no agreed standard for computational evaluation, and few papers perform user evaluations [but see~\cite{dodge2019explaining,lucic2020does}]. With respect to semi-factual explanations, there is only one relevant paper, a case-based reasoning work detailing \textit{a fortiori reasoning}~\cite{nugent2009gaining}, which follows a similar explanation paradigm to semi-factuals, but this focused only on tabular data.

The applicability of the above techniques to image data remains an open question, largely due to the difference between discrete (e.g., tabular and text) and non-discrete domains (i.e., images). In image datasets, a separate literature examines counterfactuals for adversarial attacks, rather than generating them for XAI. In adversarial attacks, small changes are made (i.e., at the pixel level of an image) to generate synthetic instances to induce misclassifications~\cite{goodfellow2014explaining}. Typically, these micro-level perturbations are constructed to be human-undetectable. In XAI however, counterfactual feature changes need to be human detectable, comprehensible, and plausible (see Fig.~\ref{Fig:1}). With this in mind, some recent work has notably used variational autoencoders (VAEs)~\cite{kingma2013auto} and generative adversarial networks (GANs)~\cite{goodfellow2014generative} to produce counterfactual images with large featural-changes for XAI. Within this literature, the most relevant research to ours are those which utilize GANs to produce explanations~\cite{samangouei2018explaingan,seah2019chest,singla2019explanation,liu2019generative}, but only one of these methods is able to offer explanations for pre-trained CNNs in multi-class classification~\cite{liu2019generative}, which we compare our method to here (see Expt.~1). This preference for binary classification is partly because choosing a counterfactual class in multi-class classification is non-trivial, and optimization to arbitrary classes is susceptible to local minima, but PIECE overcomes these issues and automates the process. In addition, none of this previous research has considered modifying exceptional features to generate explanations, or semi-factuals.

\paragraph{Present Contribution.} This paper reports PlausIble Exceptionality-based Contrastive Explanations (PIECE), a novel algorithm for generating contrastive explanations for any CNN. PIECE automatically models the distributions of learned latent features to detect \enquote{exceptional features} in a test instance, modifying them to be \enquote{normal} in explanation generation. PIECE automates the counterfactual generation process in multi-class classification, and is applicable to any pre-trained CNN. Experimental tests show that this method advances the state-of-the-art for counterfactual explanations in quantitative measurements (see Expt.~1). Additionally, semi-factual explanations are considered here for the first time in deep learning, and PIECE is shown to produce them appreciably better than other methods (see Expt.~2). So, \textit{post-hoc} explanation in XAI is significantly advanced by this work.


\section{PlausIble Exceptionality-based Contrastive Explanations (PIECE)}
\label{section:2}
\textit{Plausibility} is the major challenge facing contrastive explanations for XAI. A good counterfactual explanation needs to be plausible, informative, and actionable~\cite{poyiadzi2020face,byrne2019counterfactuals}. For example, good counterfactual explanations in a loan application system should not propose implausible feature-changes (e.g., \enquote{If you earned \$1M more, you would get the loan}). For images, plausible counterfactuals need to modify human-detectable features (see Fig.~\ref{fig:intro}); indeed, some methods can generate synthetic instances that are not even within the data distribution~\cite{laugel2019dangers}. Accordingly, an explanation-instance's proximity to the data distribution is now commonly used as a proxy for evaluating plausibility~\cite{van2019interpretable,samangouei2018explaingan}, which we use as our approach for evaluation.

Fig.~\ref{fig:intro} illustrates some of PIECE's plausible contrastive explanations for a CNN's classifications on MNIST~\cite{lecun2010mnist}, alongside a factual explanation for completeness. In Fig.~\ref{fig:intro}c, the test image of an \enquote{8} misclassified as a \enquote{3}, is shown alongside its counterfactual explanation, showing feature changes that would cause the CNN to classify it as an \enquote{8} (i.e., the cursive stroke making the plausible \enquote{8} image). An implausible counterfactual, generated by a minimal-edit method (i.e., the \textit{Min-Edit} method in Expt.~1), is also shown, with human-undetectable feature-changes that would also cause the CNN to classify the image as an \enquote{8}. Fig~\ref{fig:intro}b shows a semi-factual, with meaningful changes to the test image that do \textit{not} change the CNN's prediction. That is, \textit{even if} the \enquote{9} had a very open loop, so it looked more like a \enquote{4}, the CNN would \textit{still} classify it as a \enquote{9}. This type of explanation has potential to convince people the original classification was definitely correct~\cite{byrne2019counterfactuals,nugent2009gaining}. Finally, though these examples show two explanations for incorrect predictions (factual and counterfactual), and one for a correct prediction (semi-factual), it should be noted that these three explanation types may be generated for either predictive outcome.

PIECE uses an \textit{experience-guided} approach, exploiting the distributional properties of the training data. The algorithm generates counterfactuals and semi-factuals by identifying \enquote{exceptional} features in the test image, and then modifying these to be \enquote{normal}. This idea is inspired by people's spontaneous use of counterfactuals, specifically the \textit{exceptionality effect}, were people change exceptional events into what would \textit{normally} have occurred~\cite{byrne2019counterfactuals}. For example, when people are told that \enquote{Bill died in a car crash taking an unusual route home from work}, they typically respond counterfactually, saying \enquote{if only he had taken his \textit{normal} route home, he might have lived}~\cite{byrne2016counterfactual}. So, PIECE identifies probabilistically-low feature-values in the test image (i.e., exceptional features) and modifies them to be their expected values in the counterfactual class (i.e., normal features).

\subsection{The Algorithm: PIECE}

\begin{figure*}[!t]
\label{Fig:3}
  \centering
  \includegraphics[width=\textwidth]{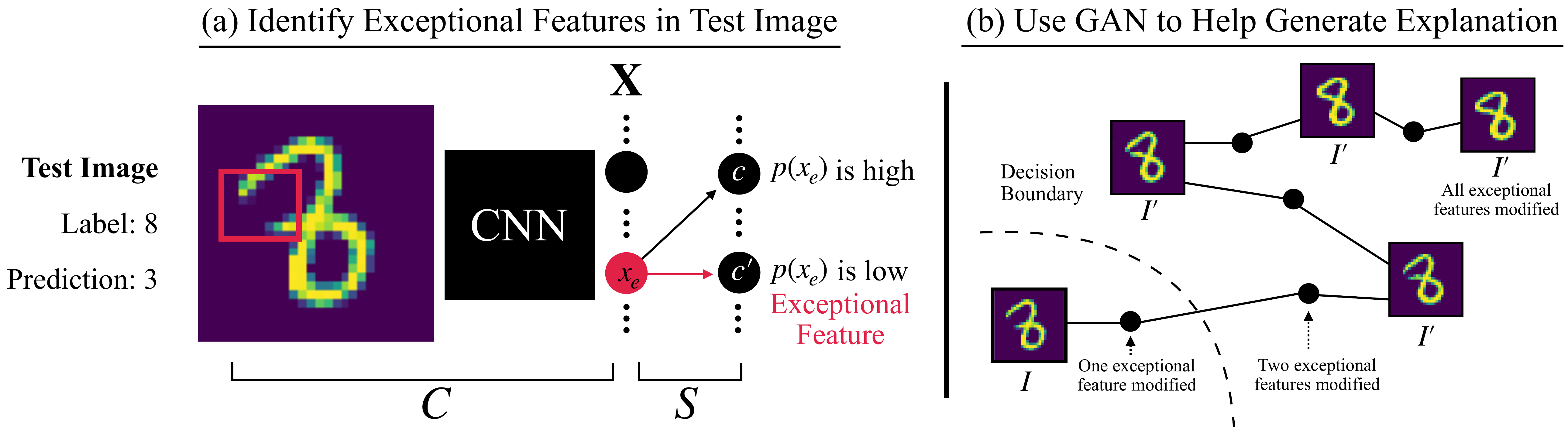} 
  \caption{PIECE Explains an \textit{Incorrect} Prediction Using a Counterfactual: The test image labelled as \enquote{8} is misclassified as a \enquote{3} by the CNN. To show how the image would have to change for the CNN to classify it as an \enquote{8}, PIECE generates a counterfactual by (a) identifying the features which have a low probability of occurrence in the counterfactual class $c'$ (i.e., the \enquote{8} class) before modifying them to be their expected feature values for $c'$, and (b) using the GAN to visualize the image $I'$ (here we show progressive exceptional-feature changes that gradually produce a plausible counterfactual image of an \enquote{8}).}
  \label{fig:algorithm}
\end{figure*}

PIECE involves two distinct systems, a CNN with predictions to be explained, and a GAN that helps generate counterfactual or semi-factual explanatory images (see Section S1 supplement for model architectures). This algorithm will work with any CNN post-training, provided there is a GAN trained on the same dataset as the CNN. PIECE has three main steps: (i) \enquote{exceptional} features are identified in the CNN for a test image from the perspective of the counterfactual class, (ii) these are then modified to be their expected values, and (iii) the resulting latent-feature representation of the explanatory counterfactual is visualized in the pixel-space with help from the GAN. To produce semi-factuals, the algorithm is identical, but the feature modifications in step two are stopped prematurely before the model's prediction crosses the counterfactual decision boundary.

\paragraph{Setup and Notation.} 
Allow all layers in the CNN up to the penultimate extracted feature layer $\mathbf{X}$ be $C$, and its output classifier $S$ (see Fig.~\ref{fig:algorithm}). The extracted features from a test image $I$ at layer $\mathbf{X}$ will be denoted as $x$, this connects to an output SoftMax layer to give a probability vector $Y$ which predicts a class $c$. To denote that $c$ is the class in $Y$ with the largest probability (i.e., the predicted class), $Y_c$ will be used. Let the generator in the GAN be $G$, and its latent input $z$, which together produce a given image. The counterfactuals to a test image $I$, in class $c$, with latent features $x$, are denoted as $I'$, $c'$ and $x'$, respectively.

\paragraph{Identify the Counterfactual Class.}
The initial steps involve locating a given test image $I$ in $G$, and then identifying the counterfactual class $c'$. First, to find the input vector $z$ for $G$, such that $G(z)\approx I$, we solve the following optimization with gradient descent:
\begin{equation}
    z = \underset{z_0}{\arg\min} \|C(G(z_0)) - C(I)\|^2_2 + \|G(z_0) - I\|^2_2 
    \label{eq:locate_image}
\end{equation}

where $z_0$ is a sample from the standard normal distribution. More efficient methods exist to do this involving encoders~\cite{seah2019chest}, but Eq.~\eqref{eq:locate_image} was sufficient here, and our focus is on more novel questions. Secondly, the counterfactual class $c'$ for $I$ may need to be generated for an incorrect or correct prediction. When the CNN incorrectly classifies $I$, $c'$ is trivially selected as being the actual label (see Fig.~\ref{fig:algorithm}). However, when the CNN's classification is correct for $I$, identifying $c'$ becomes non-trivial. We use a novel method here involving gradient \textit{ascent} to solve this problem and run:
\begin{equation}\label{eq:gradient_ascent}
    \underset{z}{\arg\max} \|S(C(G(z))) - Y_c\|^2_2 
\end{equation}

where $Y_c$ is binary encoded as all 0s, and a 1 for the class $c$. During this optimization process, the first time a decision boundary is crossed, the new class is selected as $c'$. Whilst hard-coding $c'$ can result in the optimization becoming \enquote{stuck}~\cite{liu2019generative}, our method never failed to generate the desired counterfactual, and required no human intervention.

\subsection{Step 1: Identifying Exceptional Features}
 Here, when the CNN classifies a test image $I$ as class $c$, we identify its exceptional features in $x$ by considering the statistical probability that each took its respective value, but from the perspective of $c'$. So, assuming the use of ReLU activations in $\mathbf{X}$, we can model each neuron $\mathbf{X}_i$ for $c'$, as a hurdle model with: 
\begin{equation}\label{eq:hurdle}
    p(x_i) = (1-\theta_i)\delta_{(x_i)(0)} + \theta_i f_i(x_i),   \quad     \text{s.t.} \quad x_i \geq 0
\end{equation}

where $x_i$ is the neuron activation value, $\theta_i$ is the probability of the neuron $i$ activating for the class $c'$ (i.e., Bernoulli trial success), $f_i$ is the subsequent probability density function (PDF) modelled for when $x_i > 0$ (i.e., when the \enquote{hurdle} is passed), the constraint of $x_i \geq 0$ refers to the ReLU activations, and $\delta_{(x_i)(0)}$ is the Kronecker delta function, returning 0 for $x_i>0$, and 1 for $x_i=0$. Moving forward, $X_i$ will signify the random variable associated with $f_i$.

To model this, $x$ is gathered from all training data into the latent dataset $L$, and considering the $n$ output classes, we divide $L$ into $\{L_i\}_{i=1}^{n}$ where $\forall x \in L_i, S(x) = Y_i$. Now considering the counterfactual class data $L_{c'}$, let all data for some neuron $\mathbf{X}_i$ be $\{x_j\}_{j=1}^{m} \in L_{c'}$, where $m$ is the number of instances. If we let the number of these $m$ instances where $x_j \neq 0$ be $q$, the probability of success $\theta_i$ in the Bernoulli trail can be modelled as $\theta_i=q/m$, and the probability of failure as $1-\theta_i$. The subsequent PDF $f_i$ from Eq.~\eqref{eq:hurdle} is modelled with $\{x_j\}_{j=1}^{m} \in L_{c'}, \forall x_j > 0$. Importantly, the hurdle models use what $S$ \textit{predicted} each instance to be (rather than the \textit{label}), because we wish to model what the CNN has learned, irrespective of whether it is objectively correct or incorrect.

We found empirically that the PDFs will typically approximate a Gaussian, Gamma, or Exponential distribution. Hence, we automated the modelling process by fitting the data with all three distributions (with and without a fixed location parameter of 0) using maximum likelihood estimation. Then, using the Kolmogorov-Smirnov test for goodness of fit across all these distributions, we chose the one of best fit. In all generated explanations, the average $p$-value for goodness of fit was $p$ $>$ 0.3 across all features. With the modelling process finished, a feature value $x_i$ is considered an exceptional feature $x_e$ for the test image $I$ if:
  \begin{equation}
  x_i = 0 \mid p(1 - \theta_i) < \alpha 
    \label{eq:except1}
  \end{equation} 
  \begin{equation}
  x_i > 0 \mid p(\theta_i) < \alpha 
    \label{eq:except2}
  \end{equation} 

Glossed, Eq.~\eqref{eq:except1} dictates that it is exceptional if a neuron $\mathbf{X}_i$ does not activate, given the probability of it not activating is less than $\alpha$ for $c'$ typically. Eq.\eqref{eq:except2} illustrates that it is exceptional if a neuron activates, given that the probability of it activating is less than $\alpha$ for $c'$ typically. The other two exceptional feature events are:
  \begin{equation}
  \theta_i F_i(x_i) < \alpha \mid x_i > 0
    \label{eq:except3}
  \end{equation} 
  \begin{equation}
  (1-\theta_i) + \theta_i F_i(x_i)  > 1 - \alpha \mid x_i > 0 
    \label{eq:except4}
  \end{equation}

where $F_i$ is the cumulative distribution function for $f_i$. Eq.~\eqref{eq:except3} dictates that, given the neuron has activated, it is exceptional (i.e., a probability $<$ $\alpha$) to have such a low activation value for $c'$. Eq.~\eqref{eq:except4} relays that, given the neuron has activated, it is exceptional to have such a high activation value for $c'$. In defining the $\alpha$ threshold, the statistical hypothesis-testing standard was adopted, categorizing any feature value which has a probability less than $\alpha=0.05$ as being exceptional in both experiments.

\subsection{Step 2: Changing the \textit{Exceptional} to the \textit{Expected}}
The exceptional features $\{x_e\}_{e=1}^{n} \in x$ (where $n$ is the number of exceptional features identified) divide into those that negatively or positively affect the classification of $c'$ in $I$, PIECE only modifies the former (see Algorithm \ref{algo:1}). Importantly, features are only modified if they meet the criteria regarding their connection weight, and identification process (i.e., found using Eq.~\eqref{eq:except1}/\eqref{eq:except2}/\eqref{eq:except3} or \eqref{eq:except4}). Glossed, the algorithm only modifies the exceptional feature values to their expected values if doing so brings the CNN closer to modifying the classification to $c'$. These exceptional features are ordered from the lowest probability to the highest, which is important in semi-factual explanations where the modification of features is stopped short of the decision boundary.

\IncMargin{1em}
\begin{algorithm2e}[ht]
    
    \KwIn{$x$: The latent features of the test image $I$ }
    \KwIn{$w$: The weight vector connecting $\mathbf{X}$ to $c'$}
    
    \SetArgSty{textnormal}

    \ForEach (\Comment*[f]{Ordered lowest probability to highest}) {$x_e$ in $\{x_e\}_{e=1}^{n} \in x$ }  {  
        \uIf{$w_e$ $>$ 0 \textbf{and} $x_e$ discovered with Eq.~\eqref{eq:except1}, Eq.~\eqref{eq:except2}, or Eq.~\eqref{eq:except3}} {  
            $x_e \leftarrow E[X_e]$ \DontPrintSemicolon \Comment*[r]{Using PDF modelled for $c'$ in Eq.~\eqref{eq:hurdle}}
        }\uElseIf{$w_e$ $<$ 0 \textbf{and} $x_e$ discovered with Eq.~\eqref{eq:except2} or Eq.~\eqref{eq:except4}} {
             $x_e \leftarrow E[X_e]$  \DontPrintSemicolon \Comment*[r]{Using PDF modelled for $c'$ in Eq.~\eqref{eq:hurdle}}
        }
}
\Return{$x$ (now modified to be $x'$)}
\caption{Modify exceptional features in $x$ to produce $x'$}
\label{algo:1}
\end{algorithm2e}
\DecMargin{1em}

\subsection{Step 3: Visualizing the Explanation}

Finally, having constructed $x'$, the explanation is visualized by solving the following optimization problem with gradient descent: 
\begin{equation}
    z' = \underset{z}{\arg\min} \|C(G(z)) - x'\|^2_2
    \label{eq:PIECE}
\end{equation}

and inputting $z'$ into $G$ to visualize the explanation $I'$.


\section{Experiment 1: Counterfactuals}
\label{section:3}

\begin{table*}[!t]

\label{table:1}
\centering 

\begin{tabular}{lllllllllr}

\toprule

 \multirow{2}{*}{Method}  & \multicolumn{2}{c}{MC Mean} & \multicolumn{2}{c}{MC STD} & \multicolumn{2}{c}{NN-Dist} & \multicolumn{2}{c}{IM1}  & R\%-Sub  \\

\cmidrule{2-10}

     & \# 1         & \# 2        & \# 1        & \# 2        & \# 1         & \# 2        & \# 1       & \# 2                & \#1 \\
\toprule

Min-Edit    & 0.52         &  0.61          & 0.24                & 0.13                  & 1.02                & 1.48               & 0.91                   &  1.17             & 42.87    \\
C-Min-Edit  & 0.50         &  0.45          & 0.25                & 0.14                 & 1.03                  & 1.50               & 0.93                  & 1.21              & 40.33       \\
Proto-CF    & 0.53         & N/A           & 0.23                & N/A                  & 1.02                 & N/A                & 1.28                  & N/A    & 34.75    \\
CEM         & 0.62          & N/A           & 0.22                & N/A                & 0.99                  & N/A                & 1.13                & N/A      & 43.87     \\
PIECE       & \textbf{0.99} & \textbf{0.96}  & \textbf{0.02}        &  \textbf{0.02}      & \textbf{0.41}         & \textbf{1.17}       & \textbf{0.72}       & \textbf{1.15}              & \textbf{69.32}      \\
\bottomrule

\end{tabular}

\caption{The average performance over the test-sets of the five counterfactual explanation methods for dataset \#1 (MNIST) and dataset \#2 (CIFAR-10) in Expt.~1, where the best results are highlighted in bold. R\%-Sub is tested on MNIST only.}

\label{tab:booktabs}
\end{table*}

In this experiment, PIECE's performance is compared against other known methods for counterfactual explanation generation. The tests compare PIECE against other sufficiently general methods which are applicable to color datasets~\cite{liu2019generative,wachter2017counterfactual} [here we use CIFAR-10~\cite{cifar10}], and then with the addition of other relevant works which focused on MNIST~\cite{dhurandhar2018explanations,van2019interpretable}. The methods compared in Expt.~1 are:

\begin{itemize}
  \item \textbf{PIECE.} The present algorithm, using Eq.~\eqref{eq:PIECE}, where all exceptional features were categorized with $\alpha = 0.05$, and subsequently modified.
 
  \item \textbf{Min-Edit.} A simple minimal-edit perturbation method based on a direct optimization towards $c'$, where the optimization used gradient descent and was immediately stopped when the decision boundary was crossed, defined by:  \\ 
  $ z' = \underset{z}{\arg\min} \|S(C(G(z))) - Y_{c'}\|^2_2 $. 
    
  \item \textbf{Constrained Min-Edit (C-Min-Edit).} A modified version of~\cite{liu2019generative},\footnote{They used the pixel rather than latent-space in $d(.)$. We tested both but found no significant difference. However, the latent-space required a smaller $\lambda$ to find $z'$, and was more stable~\cite{russell2019efficient}.}, and inspired by ~\cite{wachter2017counterfactual}, this optimized with gradient descent and stopped when the decision boundary was crossed, defined as: 
  \\ $ z' = \text{arg} \underset{z}{\min} \ \underset{\lambda}{\max} \quad \lambda \|S(C(G(z))) - Y_{c'}\|_2^2 + d(C(G(z)), x)$.
  
  \item \textbf{Contrastive Explanations Method (CEM).} Pertinent negatives from~\cite{dhurandhar2018explanations}, which are a form of counterfactual explanation, implemented here using \cite{alibi}. 
 
  \item \textbf{Interpretable Counterfactual Explanations Guided by Prototypes (Proto-CF).} The method by~\cite{van2019interpretable}, implemented here using \cite{alibi}. 
  
\end{itemize}

Hyperparameter choices are presented in Section S2 of the supplementary material. Although other similar techniques are reported in the literature~\cite{singla2019explanation,samangouei2018explaingan,seah2019chest}, they are not applicable as they cannot explain CNNs which are pre-trained on multi-class classification problems.

\paragraph{Setup, Test Set, and Evaluation Metrics.} For MNIST, a test-set of 163 images classified by the CNN was used which divided into: (i) correct classifications (N=60) with six examples per number-class, (ii) close-correct classifications (N=62), that had an output SoftMax probability $<$ 0.8, where the CNN \enquote{just} got the classification right,\footnote{We understand SoftMax probability is not considered reliable for CNN certainty, but it's a good baseline~\cite{hendrycks2016baseline}.} and (iii) incorrect classifications (N=41) by the CNN (i.e., every instance misclassified by the CNN). For CIFAR-10, the test-set was divided into: (i) correct classifications (N=30) with three examples per class, and (ii) incorrect classifications (N=30) with three examples per class. All instances were randomly selected, with the obvious exception of MNIST's incorrect classifications. 

Although many measures have been proposed to quantitatively evaluate an explanation's plausibility, there are no agreed benchmark measures, but most researchers use some measure of proximity to the data distribution. One related work proposed IM1 and IM2, based on training multiple autoencoders (AEs) to test the generated counterfactual's relative reconstruction error~\cite{van2019interpretable}. However, as there can be issues interpreting IM2~\cite{mahajan2019preserving}, we replaced it with Monte Carlo Dropout~\cite{gal2016dropout} (MC Dropout), a commonly used method for out-of-distribution detection~\cite{malinin2018predictive}, with 1000 forward passes. Additionally, we use R\%-Substitutability~\cite{samangouei2018explaingan} which measures how well generated explanations can substitute the actual training data. As there is relatively few explanations generated compared to the actual training datasets (163 compared to 60,000), we use $k$-NN on the pixel space of MNIST, as the classifier works well with small amounts of training data, and the centred nature of the MNIST dataset means it performs well normally (i.e., $\sim$~97\% accuracy). In the current experiment, the measures used were:

\begin{itemize}
  \item \textbf{MC-Mean.} Posterior mean of MC Dropout on the generated counterfactual image (higher is better).
  \item \textbf{MC-STD.} Posterior standard deviation of MC Dropout on the generated counterfactual (lower is better).
  \item \textbf{NN-Dist.} The distance of the counterfactual's latent representation at layer $\mathbf{X}$ from the nearest training instance measured with the $L_2$ norm [i.e., the closest \enquote{possible world}~\cite{wachter2017counterfactual}].
 \item \textbf{IM1.} From~\cite{van2019interpretable}, an AE is trained on class $c$ (i.e., $AE_c$) and $c'$ (i.e., $AE_{c'}$) to compute $\text{IM1} = \frac{\|I' - AE_{c'}(I')|\|_2^2}{\|I' - AE_{c}(I')\|_2^2}$, where a lower score is considered better.
  \item \textbf{Substitutability (R\%-Sub).} Inspired by~\cite{samangouei2018explaingan}, the method's generated counterfactuals are fit to a $k$-NN classifier (in pixel space) which predicts the MNIST test set. The original training set gives $\sim$~97\% accuracy with $k$-NN, if a method produces half that accuracy, its R\%-Sub score is 50\%.
\end{itemize}

\paragraph{Results and Discussion.} PIECE generates counterfactual explanations that are more plausible compared to the other methods in all tests, analysis using the Anderson-Darling test (AD) showed \textit{AD} $>$ 22, $p$ $<$ .001 significance to all these results (except IM1 on CIFAR-10). Notably, Proto-CF/CEM were the only methods that failed to find a counterfactual explanation for 20/25 images out of a total of 163 on MNIST, respectively. Interestingly, for all results on MNIST, a plot of the NN-Dist measure against the MC-Mean/MC-STD scores show a significant linear relationship $r$ = -0.8/0.82. So, the more a generated counterfactual is grounded in the training data, the more likely it is to be plausible [as some have argued should be the case~\cite{laugel2019dangers}], see Section S4 of the supplementary material for these plots.

\begin{figure*}[!t]
\label{Fig:4}
  \centering
  \includegraphics[width=\textwidth]{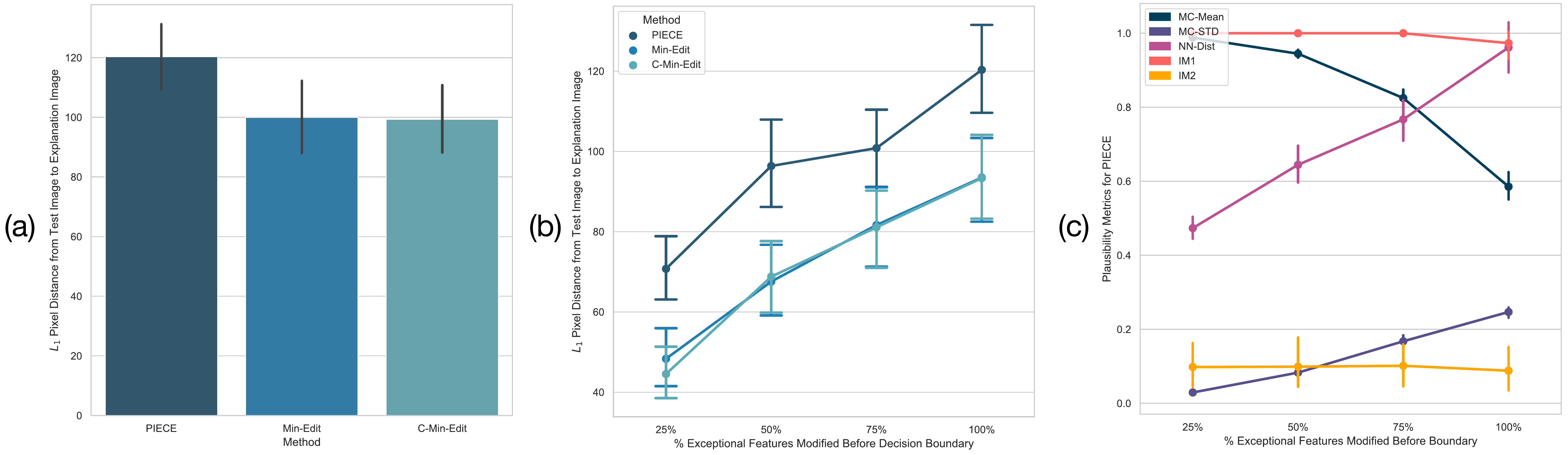} 
  \caption{Expt.~2 Results: (a) the $L_1$ pixel-space change between the test image and explanatory image from all three methods in a max-edit semi-factual, (b) the same $L_1$ metric for the three methods under progressive proportions of feature-changes, (c) the plausibility measures for PIECE, again under the same progressive proportions of feature-changes.}
  \label{fig:expt2}
\end{figure*}


\section{Experiment 2: Semi-Factuals}
One paper~\cite{nugent2009gaining} argued that semi-factual explanation (they called it \textit{a fortiori reasoning}) should involve the largest possible feature modifications without changing the classification (e.g., \enquote{\textit{Even if} you trebled your salary, you would still not get the loan}). However, they did not consider semi-factuals for image datasets, or perform controlled experiments. As such, a new evaluation method is needed to measure \enquote{good semi-factuals} in terms of how far the generated semi-factual instance is from the test instance, without crossing the decision boundary into the counterfactual class $c'$. To accomplish this in an image domain, here we use the $L_1$ distance between the test image and synthetic explanatory semi-factual in the pixel-space (n.b., the greater the distance the better the method). In the present experiment, PIECE is only compared to the minimal-edit methods from Expt.~1 (i.e., Min-Edit and C-Min-Edit), as the other methods (i.e., CEM and Proto-CF) cannot generate semi-factuals. To thoroughly evaluate all methods, three distinct tests were carried out (see Fig.~\ref{fig:expt2}). First, a max-edit run was performed on a set of test images, where each of the three methods produced their \enquote{best semi-factual}. Specifically, Min-Edit and C-Min-Edit were allowed optimize until the next step would push them over the decision boundary into the counterfactual class $c'$, and PIECE followed its normal protocol, but stopped Algorithm \ref{algo:1} when the next exceptional feature modification to $x$ would alter the CNN classification such that $S(x) \neq Y_c$. Second, the performance of the methods, on the same test set, for different proportions of feature changes were recorded. Specifically, PIECE only modifies 25\%, 50\%, 75\%, and 100\% of the exceptional-features from the first test, whilst the min-edit methods were allowed to optimize to the same distance as PIECE (measured using $L_2$ distance) in the latent-space for each of these four distances. This second test allows us to view the full spectrum of results for semi-factuals. Third, and finally, all the plausibility measures used in Expt.~1 were applied to PIECE for the same proportional-increments of changes to the exceptional features used in the second test (measured at 25\%, 50\%, 75\% and 100\%) to get a full profile of its operation. 

\paragraph{Setup, Test-Set, and Evaluation Metrics.} PIECE was run as in Expt.~1, with the counterfactual class $c'$ being selected in the same way, and with all exceptional features being identified using $\alpha = 0.05$. For full details on hyperparameter choices see Section S2 of the supplementary material. A test set of 60 test images were used (i.e., the \enquote{correct} set from MNIST in Expt.~1), with the plausibility of PIECE being evaluated using the same metrics from Expt.~1 [but we add IM2 here since it has not been tested on semi-factuals~\cite{van2019interpretable}]. The semi-factual's goodness was measured using the $L_1$ pixel distance between the test image and the semi-factual image generated, the larger this distance, the better the semi-factual.

\paragraph{Results and Discussion.} Fig.~\ref{fig:expt2} shows the results of the first comparative tests of semi-factual explanations in XAI. First, PIECE produces the best semi-factuals, with significantly higher $L_1$ distance scores than the min-edit methods (see Fig.~\ref{fig:expt2}a; \textit{AD} $>$ 2.5, $p$ $<$ .029). Second, all methods produce better semi-factuals at every distance measured (see Fig.~\ref{fig:expt2}b), but PIECE's semi-factuals are significantly better at every distance tested (\textit{AD} $>$ 3.3, $p$ $<$ .015). Third, when different plausibility measures are applied to progressive incremental changes of the exceptional features by PIECE, there are significant changes across some (i.e., MC-Mean, MC-STD, and NN-Dist), but not all measures (i.e., IM1/IM2), perhaps suggesting the former metrics are more sensitive than the latter (see Fig.~\ref{fig:expt2}c). Notably, there is a clear trade-off between plausibility (measured in MC-Dropout measures), and NN-Dist for semi-factuals, showing that as semi-factuals get better, they may sacrifice some plausibility.


\section{Conclusion}
A novel method, PlausIble Exceptionality-based Contrastive Explanations (PIECE), has been proposed that produces plausible counterfactuals to provide \textit{post-hoc} explanations for a CNN's classifications. Competitive tests have shown that PIECE adds significantly to the collection of tools currently proposed to solve this XAI problem. Future work will extend this effort to more complex image datasets. In addition, another obvious direction would be to use recent advances in text and tabular generative models (e.g., see~\cite{xu2019modeling,radford2019language}) to extend the framework into these domains, alongside pursuing semi-factual explanations more extensively, as there remains a rich, substantial, untapped research area involving them.

\section *{Ethics Statement}

As AI systems are increasingly used in human decision making (e.g., health and legal applications), there are significant issues around the fairness and accountability of these algorithms, in addition to whether or not people have reasonable grounds to trust them. One aim of explainable AI research is to create techniques and task scenarios that support people in making these fairness, accountability, and trust judgements. The present work is part of this aforementioned research effort. In providing people with counterfactual/semi-factual explanations, there is a risk of revealing \enquote{too much} about how a system operates (e.g., they potentially convey exactly how a proprietary algorithm works). Notably, the balance of this risk is more on the side of the algorithm-proprietors than on algorithm-users, which may be where we want it to be in the interests of fairness and accountability. Indeed, these methods have the potential to reveal biases in datasets and algorithms as they reveal how data is being used to make predictions (i.e., they could also be used to debug models). The psychological evidence shows that counterfactual and semi-factual explanations elicit spontaneous causal thinking in people; hence, they may have the benefit of reducing the passive use of AI technologies, enabling better human-in-the-loop systems, in which people have appropriate (rather than inappropriate) trust.

\section *{Acknowledgements}
This paper emanated from research funded by (i) Science Foundation Ireland (SFI) to the Insight Centre for Data Analytics (12/RC/2289 P2), (ii) SFI and DAFM on behalf of the Government of Ireland to the VistaMilk SFI Research Centre (16/RC/3835), and (iii) the SFI Centre for Research
Training in Machine Learning (18/CRT/6183).

\bibliography{references}

\end{document}